\documentclass[letterpaper]{article} 
\usepackage{aaai2026}  
\usepackage{times}  
\usepackage{helvet}  
\usepackage{courier}  
\usepackage[hyphens]{url}  
\usepackage{graphicx} 
\usepackage{makecell}
\usepackage{array}
\usepackage{enumitem}
\usepackage{multirow}
\usepackage{subcaption}
\usepackage{graphicx}
\usepackage{tabularx}
\usepackage{color}
\usepackage{xspace} 
\usepackage[utf8]{inputenc}
\usepackage[T1]{fontenc}
\usepackage{enumitem}
\usepackage{diagbox}
\usepackage[flushleft]{threeparttable}
\usepackage{marvosym}
\usepackage{amssymb}
\usepackage{amsmath}
\usepackage{booktabs}
\usepackage{amsthm}

\usepackage{natbib}  
\usepackage{caption} 
\usepackage{algorithm}
\usepackage{algorithmicx}
\usepackage{algpseudocode}
\usepackage{amsmath}
\usepackage{amssymb}
\nocopyright
\usepackage{newfloat}
\usepackage{listings}
\DeclareCaptionStyle{ruled}{labelfont=normalfont,labelsep=colon,strut=off} 
\floatstyle{ruled}
\newfloat{listing}{tb}{lst}{}
\floatname{listing}{Listing}
\title{Rethinking Irregular Time Series Forecasting: A
Simple yet Effective Baseline}

\author{
    Xvyuan Liu\equalcontrib$^{1}$, 
    Xiangfei Qiu\equalcontrib$^{1}$, 
    Xingjian Wu\equalcontrib$^{1}$, 
    Zhengyu Li$^{1}$, \\
    \textbf{Chenjuan Guo$^{1}$, 
    Jilin Hu$^{1,2}$\thanks{Corresponding Author.}, 
    Bin Yang$^{1}$}
}
\affiliations{
    \textsuperscript{\rm 1}School of Data Science and Engineering, East China Normal University, Shanghai, China\\
    \textsuperscript{\rm 2}Engineering Research Center of Blockchain Data Management, Ministry of Education, China\\
    \{xvyuanliu, xfqiu, xjwu, lizhengyu\}@stu.ecnu.edu.cn, \{cjguo, jlhu, byang\}@dase.ecnu.edu.cn
}

\usepackage{bibentry}

\begin{document}

\maketitle

\begin{abstract} 
The forecasting of irregular multivariate time series (IMTS) is crucial in key areas such as healthcare, biomechanics, climate science, and astronomy. However, achieving accurate and practical predictions is challenging due to two main factors. First, the inherent irregularity and data missingness in irregular time series make modeling difficult. Second, most existing methods are typically complex and resource-intensive. In this study, we propose a general framework called APN to address these challenges. Specifically, we design a novel Time-Aware Patch Aggregation (TAPA) module that achieves adaptive patching. By learning dynamically adjustable patch boundaries and a time-aware weighted averaging strategy, TAPA transforms the original irregular sequences into high-quality, regularized representations in a channel-independent manner. Additionally, we use a simple query module to effectively integrate historical information while maintaining the model's efficiency. Finally, predictions are made by a shallow MLP. Experimental results on multiple real-world datasets show that APN outperforms existing state-of-the-art methods in both efficiency and accuracy.
\end{abstract}

\section{Introduction}
\label{introduction}

Irregular Multivariate Time Series (IMTS) data are widely observed in various domains such as healthcare, biomechanics, climate science, and astronomy~\citep{Yao2018, GRU-ODE, Neural-CDE, qiu2025DBLoss,qiu2025duet, wu2025k2vae,gao2025ssdts,hu2024multirc}. Irregular Multivariate Time Series Forecasting (IMTSF) is a crucial research task that provides valuable insights for early warning and proactive decision-making. However, the inherent irregularity of observations and missing data~\citep{GraFITi, tPatchGNN} in IMTS poses significant challenges to IMTSF modeling.

\begin{figure}[t]  
  \centering  
  \raisebox{0pt}[\height][\depth]{\includegraphics[width=1.0\columnwidth]{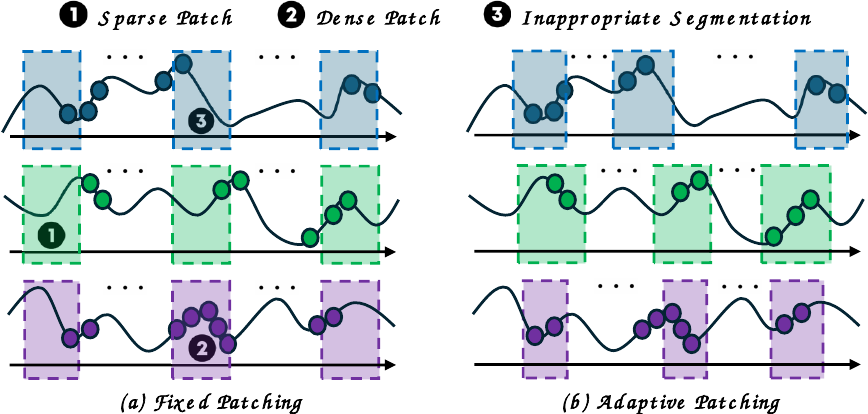}}
  \caption{Fixed Patching vs. Adaptive Patching.}  
  \label{fixed_span_patching} 
\end{figure}

To address this challenge, recent IMTSF methods, such as tPatchGNN \citep{tPatchGNN} and TimeCHEAT \citep{TimeCHEAT}, have adopted the Fixed Patching approach---see Figure~\ref{fixed_span_patching}a. This approach divides the time series into fixed-length patches with equal intervals among them. However, this approach has notable limitations: 1) \textit{Uneven Information Density Across Patches:} Fixed Patching struggles to adapt to local variations in data density, leading to uneven information density among patches. For example, sparse patches~(where the number of observations is limited) may result in insufficient feature extraction, yet dense patches~(where the number of observations is abundant) may contain redundant information or noise, thereby impairing the extracted features. \textit{2) Inappropriate Segmentation of Key Semantic Information:} Fixed Patching risks splitting critical dynamic information, which hampers the model's ability to capture the complete semantic context. Therefore, \textbf{the first challenge is how to design an adaptive patching approach} that can adapt to the local information density variations of irregular multivariate time series and capture complete semantic information.

Meanwhile, existing IMTSF models generally suffer from high computational costs and long running times. For example, neural-ODE-based models \citep{Neural-ODE, Latent-ODE, TG-ODE} require computationally intensive numerical solvers to accurately model continuous-time dynamics \citep{mTAN, Neural-ODE}. GNN-based models \citep{GraFITi, tPatchGNN} are hindered by the overhead of complex graph construction and multi-round node information aggregation \citep{BadGNN}; Transformer-based models \citep{ContiFormer, Warpformer}, which utilize multi-layer self-attention mechanisms and feedforward networks, often result in large parameter scales \citep{BadTransformer}. These models typically construct computationally intensive and parameter-heavy complex architectures to effectively handle the intricate dependencies and dynamic changes in IMTS. While this design enhances the model's performance to some extent, it also incurs high computational costs and long runtimes, limiting its practical application in resource-constrained scenarios. Notably, in regular time series forecasting tasks, models such as SparseTSF \citep{lin2024sparsetsf}, and CycleNet \citep{lincyclenet} have significantly reduced computational overhead and achieved competitive prediction accuracy by adopting simple architectures with fewer parameters. Therefore, \textbf{the second challenge is how to design an efficient model for IMTSF.}

To address the above challenges, we propose a general framework called APN. Specifically, we design a novel Time-Aware Patch Aggregation (TAPA) module that achieves adaptive patching---see Figure~\ref{fixed_span_patching}b. By learning dynamically adjustable patch boundaries and a time-aware weighted averaging strategy, TAPA transforms the original irregular sequences into high-quality, regularized representations in a channel-independent manner. Based on the representations, we use a simple query module to effectively integrate historical information while maintaining the model's efficiency. Finally, predictions are made by a shallow MLP. Results on multiple real-world datasets show that APN outperforms existing SOTA methods in both efficiency and accuracy.

The contributions of our paper are summarized as follows:
\begin{itemize}[left=0.3cm]
 \item To address IMTSF, we propose a general framework named APN. This framework leverages adaptive patching to generate high-quality and regular initial patch representations. Based on these representations, we employ a simple query module to integrate contextual information, ensuring the effective design of the framework.
 \item We design a novel TAPA module to achieve adaptive patching. By learning dynamically adjustable patch boundaries and a time-aware weighted averaging strategy, TAPA transforms the original irregular sequences into high-quality, regularized
representations in a channel-independent manner, effectively adapting to local variations in information density and capturing complete semantics.
 \item We conduct experiments on multiple datasets. The results show that APN outperforms existing SOTA baselines in both forecasting accuracy and computational efficiency. Additionally, all datasets and code are available at: \url{https://github.com/decisionintelligence/APN}.
\end{itemize}

\section{Related Work}
\label{sec:related_work}

\subsection{Progress in Irregular Multivariate Time Series Forecasting}
\label{subsec:imts_forecasting_methods}
IMTSF is crucial for key domains such as healthcare and climate science. The inherent characteristics of IMTS, such as non-uniform sampling intervals and asynchronous channels, present significant challenges to regular time series forecasting models. To address these characteristics, researchers have proposed various IMTSF models. Some models employ approaches based on continuous-time dynamics \citep{Neural-ODE, Latent-ODE, CRU, GRU-ODE}, utilizing ordinary or stochastic differential equations (ODE/SDE) to adapt to irregular sampling points. For instance, Neural Flows \citep{Neural-Flows} proposes directly modeling the solution curves of ODEs, thereby avoiding the costly numerical integration steps in traditional ODE solvers. GRU-ODE-Bayes \citep{GRU-ODE} innovatively combines the idea of Gated Recurrent Units (GRU) with ODEs and effectively handles sparse observational data through a Bayesian update mechanism. Other models leverage graph neural networks and attention mechanisms \citep{GraFITi, tPatchGNN, TimeCHEAT} to capture complex dependencies in IMTS. For example, GraFITi \citep{GraFITi} transforms IMTS into sparse bipartite graphs and predicts edge weights through GNNs. tPatchGNN \citep{tPatchGNN} innovatively segments irregular sequences into time-aligned patches and combines Transformer and adaptive GNNs to handle intra-patch and inter-patch dependencies, respectively.

\subsection{Progress in Patch-based Irregular Multivariate Time Series Forecasting}
\label{subsec:patching_progress_imts}
The patch-based strategy, which has proven successful for regular time series \citep{PatchTST, Crossformer,wu2025enhancing,qiu2025comprehensive,qiu2025dag,lu2025dtaf,wang2025plug}, has been adapted for IMTS Forecasting (IMTSF). Initial adaptations, such as tPatchGNN \citep{tPatchGNN} and TimeCHEAT \citep{TimeCHEAT}, employed a fixed-span patching strategy. However, this rigid segmentation is ill-suited for the non-uniform data distribution of IMTS. It often creates patches with highly variable information content—some being too sparse for meaningful feature extraction, while others may be overly dense with redundant data.

To overcome this, adaptive patching emerged as a natural evolution. One prominent approach, primarily explored in the context of regular time series (e.g., HDMixer \citep{HDMixer}), involves learning flexible patch boundaries and then generating regularized patch representations through interpolation-based resampling. While effective for dense, regular data, this methodology is fundamentally flawed for IMTS. Interpolating across sparse or missing intervals can fabricate misleading data points, introducing significant artifacts and undermining model reliability.

In contrast, our work proposes a distinct aggregation-based adaptive patching paradigm, specifically designed for the challenges of IMTS. Instead of creating new, artificial data points, our TAPA module learns dynamic ``soft windows'' and computes each patch representation by performing a direct, weighted aggregation of all original observations. This strategy provides two critical advantages for IMTS: (1) Data Fidelity: It exclusively uses the original, observed data, thereby avoiding the distortions and potential inaccuracies of interpolation. (2) Complete Information Coverage: The soft-weighting mechanism ensures that every data point contributes to the representation of relevant patches, mathematically precluding the information loss that can occur when observations fall between the hard boundaries of other methods.

\section{Methodology}
\label{sec:methodology}
\begin{figure*}[!t]
    \centering
    \includegraphics[width=0.95\linewidth]{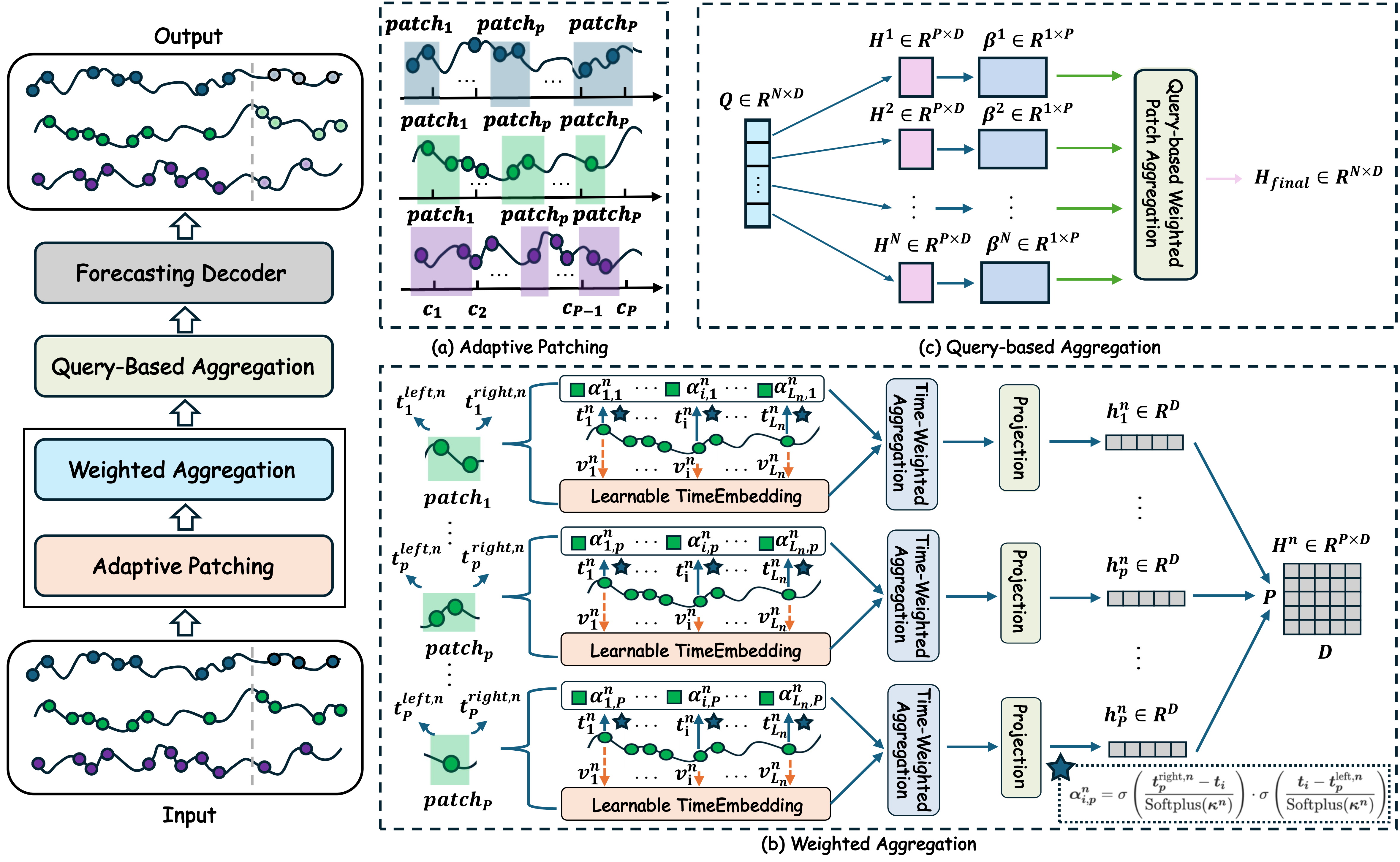}
    \caption{The overall framework of APN, which initially divides each univariate irregular time series into a series of unfixed patches using the \textit{Adaptive Patching} Module. Then the \textit{Weighted Aggregation} Module generates high-quality and regular initial patch representations. Based on the representations, the \textit{Query-based Aggregation} Module is utilized to incorporate contextual information. Finally, the \textit{Forecasting Decoder} outputs the final forecasting results. The \textit{Adaptive Patching} Module and \textit{Weighted Aggregation} Module collectively form the \textit{Time-Aware Patch Aggregation} Module.}
\label{fig:overview}
\end{figure*}

\subsection{Problem Definition}
\label{sec:problem_definition}
An Irregular Multivariate Time Series (IMTS) $\mathcal{O}$ consists of $N$ univariate sequences $\{o^n_{1:L_n}\}_{n=1}^N$. Each sequence $o^n_{1:L_n}$ comprises $L_n$ observations $(t_i^n, v_i^n)$, where the intervals between timestamps $t_i^n$ are irregular, and sampling across different variables is typically asynchronous. The IMTS forecasting task is defined as: given historical observations $\mathcal{O}$ and a set of prediction queries for all variables $\mathcal{Q} = \{[q_j^n]_{j=1}^{Q_n}\}_{n=1}^N$, construct and optimize a prediction model $\mathcal{F}(\cdot)$ capable of accurately predicting future observed values $\hat{\mathcal{V}} = \{[\hat{v}_j^n]_{j=1}^{Q_n}\}_{n=1}^N$ corresponding to each query $q_j^n$, i.e.,
\begin{equation}
\mathcal{F}(\mathcal{O}, \mathcal{Q}) \rightarrow \hat{\mathcal{V}}
\end{equation}

\subsection{Framework Overview}
\label{sec:model_overview}
The architectural design of APN is guided by a core principle: decoupling the challenge of handling irregularity from the task of forecasting. As illustrated in Figure \ref{fig:overview}, instead of relying on a monolithic, complex model, APN adopts a strategic two-stage pipeline.

The cornerstone of this pipeline is the \textit{Time-Aware Patch Aggregation} (TAPA) module (Section~\ref{sec:tapa_detail}). Operating in a channel-independent manner, its sole yet critical purpose is to transform the raw, irregular observations into a high-quality, regularized sequence of patch representations. Through its novel use of \textit{Adaptive Patching} and \textit{Weighted Aggregation}, TAPA robustly adapts to local information density and captures complete semantic units. Crucially, it achieves this without resorting to interpolation, thereby preserving data fidelity and sidestepping the introduction of artificial artifacts.

The success of this initial transformation is pivotal, as it enables the subsequent use of a remarkably efficient and lightweight architecture. A concise \textit{Query-based Aggregation} module (Section~\ref{sec:query_weighted_aggregation}) then effectively summarizes the historical context from these regularized patches for each variable. Finally, this compact representation is fed into a simple MLP-based \textit{Forecasting Decoder} (Section~\ref{sec:prediction_decoding}) for the final prediction.

In essence, APN strategically front-loads the complexity of handling irregularity into the novel TAPA module. By producing information-rich, regularized representations upfront, it obviates the need for computationally expensive back-end models, achieving state-of-the-art performance through an elegant and efficient design.

\subsection{Time-Aware Patch Aggregation (TAPA)}
\label{sec:tapa_detail}
At the heart of APN lies the Time-Aware Patch Aggregation (TAPA) module, a novel mechanism designed to fundamentally reframe how we process irregular time series. Traditional methods rely on a paradigm of ``hard segmentation,'' imposing rigid, fixed-span boundaries that are ill-suited to the non-uniform nature of IMTS. In stark contrast, TAPA introduces a ``soft aggregation'' paradigm. Instead of discretely assigning observations to patches, it conceptualizes each patch as a soft, overlapping field of influence, allowing it to aggregate information from all observations in a temporally aware manner.

This paradigm shift is realized through a cohesive, two-step process executed independently for each channel $o^n_{1:L_n} = \{(t^n_i, v^n_i)\}_{i=1}^{L_n}$. First, in Adaptive Patching, the model learns the dynamic temporal characteristics—the center and scale—of each patch's field of influence. Second, in Weighted Aggregation, it computes a representation for each patch by performing a direct, weighted aggregation of all raw observations, where the weights are determined by each observation's temporal relevance to the patch's learned characteristics. This entire process transforms the irregular input sequence into a regularized, information-rich sequence of patch representations $H^n = [h^n_1, \dots, h^n_{P}]$.

\subsubsection{Adaptive Patching: Learning Dynamic Temporal Windows}
\label{sec:adaptive_boundaries}
The first step is to define the temporal scope of each patch. Instead of imposing predefined, rigid boundaries, TAPA learns a dynamic temporal window, $[t_p^{left, n}, t_p^{right, n}]$, for each target patch $p$. This dynamism is the core mechanism by which TAPA adapts to the non-uniform distribution of data. For instance, in a sparse but critical region, the model can learn to expand a window's width to ensure sparse yet vital signals are encompassed. Conversely, in a dense region with redundant data, a window can be strategically narrowed to focus on the most salient signals, thus preventing dilution from less informative points.

This adaptive process empowers the model to carve out semantically coherent segments from the raw time series, based on the data's inherent structure. To achieve this, for each patch $p$ and channel $n$, TAPA learns two key parameters: a positional adjustment $\delta_p^{n}$ and a log-scale width parameter $\lambda_p^{n}$. The window boundaries are then computed as:
\begin{align}
    t_p^{left, n} &= c_p - \frac{S_{init}}{2} + \delta_p^{n} \label{eq:left_boundary} \\
    t_p^{right, n} &= t_p^{left, n} + \exp(\lambda_p^{n}),\label{eq:right_boundary}
\end{align}
where $T_{obs}$ is the time span of the historical observation window, with $c_p = (p - 0.5) \cdot (T_{obs}/P)$ and $S_{init} = T_{obs}/P$ representing the initial reference center and width, respectively. The use of $\exp(\lambda_p^{n})$ is a deliberate design choice, ensuring the learned window width remains strictly positive throughout optimization. This dynamic windowing mechanism allows the model to adaptively frame regions of interest, laying the foundation for a more meaningful and robust aggregation in the subsequent step.

\subsubsection{Weighted Aggregation: From Raw Observations to Rich Representations}
\label{sec:weighted_aggregation}
With the dynamic temporal windows defined, the next step is to generate patch representations by aggregating information from the raw observations.

Enriching Observations with Temporal Context. To fully leverage temporal information, we first enrich each raw observation. A learnable time embedding $TE(t_i^n) \in \mathbb{R}^{D_{te}}$ is generated for each timestamp $t_i^n$ in channel $n$. This embedding, composed of linear layers for scale and sine-activated layers for periodicity, captures both the absolute position and periodic patterns in time. 
It is then concatenated with the original value $v_i^n$ to form an augmented representation:
\begin{equation}
\tilde{v}_i^n = \text{Concat}(v_i^n, TE(t_i^n))\label{eq:value_with_time}
\end{equation}

The Soft Window Function. We then employ a time-aware soft window function to calculate the relevance weight, $\alpha_{i, p}^{n}$, of each observation $i$ to each patch $p$. The function is elegantly constructed as the product of two Sigmoid functions:
\begin{equation}
\alpha_{i, p}^{n} = \sigma\left(\frac{t_p^{right, n} - t_i^n}{\text{Softplus}(\kappa^n)}\right) \cdot \sigma\left(\frac{t_i^n - t_p^{left, n}}{\text{Softplus}(\kappa^n)}\right)\label{eq:alpha_weight}
\end{equation}
Here, the first Sigmoid term models a smooth ``falloff'' from the right boundary, while the second models a smooth ``rise'' from the left boundary. Their product creates a continuous, bell-shaped weighting curve centered within the patch's learned range. The term $\text{Softplus}(\kappa^n)$ is a learnable, strictly positive temperature that controls the softness of the window edges. A smaller temperature leads to sharper, more defined boundaries, while a larger temperature creates gentler, more overlapping fields of influence, granting the model further adaptive flexibility.

A Crucial Design Choice for Information Integrity. The choice of this soft window function is critical. Since the Sigmoid function $\sigma(x)$ is strictly positive for all real inputs, the resulting weight $\alpha_{i, p}^{n}$ is also guaranteed to be positive for any observation $i$ and any patch $p$. This property is not a mere side effect; it is a fundamental design principle that mathematically guarantees complete information coverage. Every observation contributes to the representation of every patch (albeit with varying degrees of influence), inherently preventing the information loss that plagues all hard-segmentation methods, where observations can be inadvertently discarded.

Final Aggregation. The final representation for the $p$-th patch, $\bar{h}_p^{n}$, is computed as a normalized weighted average of all augmented observation features:
\begin{equation}
\bar{h}_p^{n} = \frac{\sum_{i=1}^{L_n} \alpha_{i, p}^{n} \cdot \tilde{v}^n_i}{\sum_{i=1}^{L_n} \alpha_{i, p}^{n} + \epsilon} \in \mathbb{R}^{1+D_{te}}, \label{eq:weighted_avg}
\end{equation}
where the summation spans all $L_n$ observations, and $\epsilon$ ensures numerical stability. Finally, to enhance expressiveness and align dimensions, we project $\bar{h}_p^{n}$ into the model's uniform hidden space via a linear layer: $h_p^{n} = \text{Linear}_{D}(\bar{h}_p^{n})$. Through this entire process, each univariate irregular sequence is transformed into a refined, structurally regularized sequence $H^n = [h^n_1, \dots, h^n_{P}] \in \mathbb{R}^{P \times D}$.

\subsection{Query-based Aggregation}
\label{sec:query_weighted_aggregation}
The TAPA module delivers a sequence of high-quality, regularized patch representations $H^n = [h^n_1, \dots, h^n_{P}]$. This successfully concludes the first stage of our pipeline: handling irregularity. The second stage, contextualization and forecasting, can now proceed with remarkable efficiency. To this end, we employ a concise query-based aggregation mechanism to distill the entire historical sequence into a single, potent context vector $H_c^n$.

First, to preserve the sequential order of the patches, standard positional encodings ($PE$) are added, yielding position-aware representations $H_{pe}^n = H^n + PE$. Then, instead of resorting to complex inter-patch interactions like multi-head self-attention, we introduce a single learnable query vector $q^n \in \mathbb{R}^D$ for each channel. This query acts as a task-specific lens, dynamically assessing the importance of each patch for the final forecast. The importance scores $s_p^n$ are computed via a simple dot product, normalized into weights $\beta^n$ via Softmax, and used to compute the final context vector $H_c^n$ as a weighted sum:
\begin{align}
    s_p^n &= \frac{q^n \cdot (h_{pe,p}^n)^T}{\sqrt{D}} \quad \text{for } p=1, \dots, P \label{eq:score_calc_query_condensed} \\
    \beta^n &= \text{Softmax}(\{s_p^n\}_{p=1}^P) \label{eq:softmax_weights_query_condensed} \\
    H_c^n &= \sum_{p=1}^{P} \beta_p^n h_{pe,p}^n \label{eq:context_vector_query_condensed}
\end{align}
The resulting summary representations for all channels, $[H_c^1, \dots, H_c^N]$, form the final representation matrix $H_{c} \in \mathbb{R}^{N \times D}$. A layer normalization is then applied to stabilize the input for the decoder. This lightweight aggregation mechanism is a direct testament to our design philosophy: by front-loading the complexity into TAPA, the subsequent modules can be elegantly simple yet powerful.

\subsection{Forecasting Decoder}
\label{sec:prediction_decoding}
The final step of the APN framework is the ultimate demonstration of its architectural elegance. Having distilled the entire irregular history into a powerful, fixed-size representation $H_{c}^n$, the forecasting task becomes remarkably straightforward.

The prediction for a query time $q_k^n$ is made by a simple two-layer MLP decoder. This decoder takes the concatenated summary representation $H_{c}^n$ and the learnable temporal encoding of the query time, $TE(q_k^n)$, as input to produce the final value $\hat{v}^n_k$:
\begin{equation}
\hat{v}^n_k = \text{MLP}(\text{Concat}(H_{c}^n, TE(q_k^n))) \in \mathbb{R}
\end{equation}
The ability to use such a simple decoder is not a limitation but a feature, underscoring the richness of the representation crafted by TAPA and the query aggregator.

The model is trained end-to-end by minimizing the standard Mean Squared Error (MSE) loss between the predicted values $\hat{v}^n_k$ and the ground truth values $v^n_k$ across all channels and query points:
\begin{equation}
\mathcal{L} = \frac{1}{\sum_{n=1}^{N} Q_n} \sum_{n=1}^{N} \sum_{k=1}^{Q_n} (\hat{v}^n_k - v^n_k)^2,
\end{equation}
where $Q_n$ is the number of query points for each channel.

\section{Experiments}
\label{sec:experiments}

\begin{table}[t]
\centering
\resizebox{1\linewidth}{!}{%
\begin{tabular}{lccccc}
\toprule
\textbf{Dataset} & \textbf{\# Variables} & \textbf{\# samples} & \textbf{Avg \# Obs.} & \textbf{Max Length}  \\
\midrule
PhysioNet        & 36 & 11,981 & 308.6 & 47 \\
MIMIC            & 96 & 21,250 & 144.6 & 96 \\
HumanActivity    & 12 & 1,359 & 362.2 & 131 \\
USHCN            & 5 & 1,114 & 313.5 & 337 \\
\bottomrule
\end{tabular}}
\caption{Dataset statistics.}
\label{tab:dataset_statistics}
\end{table}

\begin{table*}[t]
    \centering
    \renewcommand{\arraystretch}{1.5}
    \resizebox{\textwidth}{!}{%
    \begin{tabular}{l|cc|cc|cc|cc}
        \toprule 
        \textbf{Dataset} & \multicolumn{2}{c|}{\textbf{HumanActivity}} & \multicolumn{2}{c|}{\textbf{USHCN}} & \multicolumn{2}{c|}{\textbf{PhysioNet}} & \multicolumn{2}{c}{\textbf{MIMIC}} \\ 
        \midrule 
        \textbf{Metric}  & \textbf{MSE}           & \textbf{MAE}           & \textbf{MSE}         & \textbf{MAE}         & \textbf{MSE}         & \textbf{MAE}         & \textbf{MSE}         & \textbf{MAE}         \\ 
        \midrule 
        PrimeNet    & 4.2507±0.0041 & 1.7018±0.0011 & 0.4930±0.0015 & 0.4954±0.0018 & 0.7953±0.0000 & 0.6859±0.0001 & 0.9073±0.0001 & 0.6614±0.0001 \\
        NeuralFlows & 0.1722±0.0090 & 0.3150±0.0094 & 0.2087±0.0258 & 0.3157±0.0187 & 0.4056±0.0033 & 0.4466±0.0027 & 0.6085±0.0101 & 0.5306±0.0066 \\
        CRU         & 0.1387±0.0073 & 0.2607±0.0092 & 0.2168±0.0162 & 0.3180±0.0248 & 0.6179±0.0045 & 0.5778±0.0031 & 0.5895±0.0092 & 0.5151±0.0048 \\
        mTAN        & 0.0993±0.0026 & 0.2219±0.0047 & 0.5561±0.2020 & 0.5015±0.0968 & 0.3809±0.0043 & 0.4291±0.0035 & 0.9408±0.1126 & 0.6755±0.0459 \\
        SeFT        & 1.3786±0.0024 & 0.9762±0.0007 & 0.3345±0.0022 & 0.4083±0.0084 & 0.7721±0.0021 & 0.6760±0.0029 & 0.9230±0.0015 & 0.6628±0.0008 \\ 
        GNeuralFlow & 0.3936±0.1585 & 0.4541±0.0841 & 0.2205±0.0421 & 0.3286±0.0412 & 0.8207±0.0310 & 0.6759±0.0100 & 0.8957±0.0209 & 0.6450±0.0072 \\
        GRU-D       & 0.1893±0.0627 & 0.3253±0.0485 & 0.2097±0.0493 & 0.3045±0.0305 & 0.3419±0.0029 & 0.3992±0.0011 & 0.4759±0.0100 & 0.4526±0.0055 \\
        Raindrop    & 0.0916±0.0072 & 0.2114±0.0072 & 0.2035±0.0336 & 0.3029±0.0264 & 0.3478±0.0019 & 0.4044±0.0020 & 0.6754±0.1829 & 0.5444±0.0868 \\
        Warpformer  & 0.0449±0.0010 & 0.1228±0.0018 & 0.1888±0.0598 & 0.2939±0.0591 & \textbf{0.3056±0.0011} & 0.3661±0.0016 & \underline{0.4302±0.0035} & \underline{0.4025±0.0014} \\ 
        tPatchGNN   & 0.0443±0.0009 & 0.1247±0.0031 & 0.1885±0.0403 & 0.3084±0.0479 & 0.3133±0.0053 & 0.3697±0.0049 & 0.4431±0.0115 & 0.4077±0.0088 \\
        GraFITi     & \underline{0.0437±0.0005} & \underline{0.1221±0.0017} & \underline{0.1691±0.0093} & 0.2777±0.0248 & \underline{0.3075±0.0015} & \textbf{0.3637±0.0036} & 0.4359±0.0455 & 0.4142±0.0297 \\ 
        \midrule 
        \textbf{APN (Ours)}       & \textbf{0.0421±0.0001} & \textbf{0.1159±0.0006} & \textbf{0.1590±0.0137} & \textbf{0.2611±0.0167} & 0.3093±0.0011 & \underline{0.3650±0.0026} & \textbf{0.4292±0.0027} & \textbf{0.4016±0.0016} \\ 
        \bottomrule 
    \end{tabular}%
    }
    \caption{Forecasting performance on four IMTS datasets. Overall performance is evaluated by MSE and MAE (mean ± std). The best and second-best results are highlighted in \textbf{bold} and with an \underline{underline}, respectively.}
    \label{tab:overall_performance_adjusted}
\end{table*}

\subsection{Setup}
\label{sec:setup}

\textbf{Datasets:} To evaluate the model's performance, we select four widely used IMTS datasets, including PhysioNet, MIMIC, HumanActivity, and USHCN. These datasets span multiple domains such as healthcare, biomechanics, and climate science. Table \ref{tab:dataset_statistics} summarizes the key statistical features of these datasets. The PhysioNet Challenge 2012 dataset provides clinical time series from the first 48 hours of ICU stays. MIMIC is a large database containing de-identified health data from ICU patients. The HumanActivity dataset consists of biomechanics data with 3D positional variables captured from subjects performing various activities. For climate science, the USHCN dataset includes historical meteorological data from stations across the United States. All datasets are partitioned into training, validation, and test sets using a standard 80\%, 10\%, and 10\% ratio, respectively.

\noindent
\textbf{Baselines:} To evaluate the performance of APN, we compare it with eleven baseline models. These baseline models can be broadly categorized into two groups: (1) IMTS Classification/Imputation Models: including PrimeNet \citep{PrimeNet}, SeFT \citep{SeFT}, mTAN \citep{mTAN}, GRU-D \citep{GRU_D}, Raindrop \citep{RainDrop}, and Warpformer \citep{Warpformer}. (2) IMTS Forecasting Models: including NeuralFlows \citep{NeuralFlows}, CRU \citep{CRU}, GNeuralFlow \citep{GNeuralFlow}, tPatchGNN \citep{tPatchGNN}, GraFITi \citep{GraFITi}.

\noindent
\textbf{Implementation Details:} All our experiments were conducted on a server equipped with an NVIDIA A800 GPU and implemented using the PyTorch 2.6.0+cu124 framework. All models are trained using the Mean Squared Error (MSE) as the loss function and optimized with the AdamW optimizer. We set the maximum number of training epochs to 200 and employ an early stopping strategy, where training is terminated if the model's performance on the validation set does not improve for 10 consecutive epochs. To ensure a fair comparison, we primarily adopted the hyperparameter settings reported in the original papers for the baseline models. Building on these configurations, we conducted a further comprehensive search and fine-tuning of key hyperparameters on the validation set for all models, including our proposed APN, to ensure each achieved a competitive level of performance. To ensure reproducibility and mitigate the effects of randomness, each experiment is run independently with five different random seeds (from 2024 to 2028), and we report the mean and standard deviation. Detailed hyperparameter configurations for all models are provided in our code repository. We do not apply the ``Drop Last'' trick~\citep{qiu2024tfb,qiu2025tab} to ensure a fair comparison.

\subsection{Main results}
\label{sec:main_results}
We compare the APN model with eleven baseline models on four challenging datasets---see Table~\ref{tab:overall_performance_adjusted}. We have the following observations: 1) APN achieves leading prediction accuracy across the board. On all datasets, APN achieves optimal or highly competitive forecasting performance. Compared to the second-best performing model, GraFITi, APN achieves significant reductions in MSE and MAE metrics by approximately 2.64\% and 3.61\%, respectively. 2) APN demonstrates exceptional cross-domain generalization capability and robustness. Whether on IMTS datasets with different characteristics in healthcare (PhysioNet, MIMIC), biomechanics (HumanActivity), or climate science (USHCN), APN consistently performs excellently. The outstanding performance of APN can be attributed to its TAPA module, which generates superior patch representations by directly performing a soft-aggregation over raw observations within adaptively learned temporal windows. This approach effectively captures salient local dynamics without the potential distortion from interpolation. Building on these high-quality representations, the streamlined query module and MLP decoder ensure a lightweight yet powerful framework.

\begin{table}[t]
    \centering
    \fontsize{7pt}{7pt}\selectfont\rmfamily
    \setlength{\tabcolsep}{3mm} 
    \begin{tabular}{c|c|c}
        \toprule
        \textbf{Method} & \textbf{MIMIC} & \textbf{USHCN} \\
        \cmidrule{1-3}
        \textbf{APN (Ours)} & \textbf{0.4292 $\pm$ 0.0027} & \textbf{0.1590 $\pm$ 0.0137} \\
        \cmidrule{1-3}
        \begin{tabular}{@{}c@{}}w/o Adaptive\\ Patching\end{tabular} & 0.4309 $\pm$ 0.0048 & 0.1717 $\pm$ 0.0283 \\
        \cmidrule{1-3}
        \begin{tabular}{@{}c@{}}w/o Weighted \\Aggregation\end{tabular} & 0.4388 $\pm$ 0.0086 & 0.1863 $\pm$ 0.0486 \\
        \cmidrule{1-3}
        \begin{tabular}{@{}c@{}}w/o Query-based\\ Aggregation\end{tabular} & 0.4672 $\pm$ 0.0051 & 0.1981 $\pm$ 0.0327 \\
        \bottomrule
    \end{tabular}
    \caption{Ablation studies for APN (MSE). Results are reported as mean $\pm$ std. Best results are highlighted in \textbf{bold}.}
    \label{tab:ablation_results}
\end{table}

\subsection{Ablation Studies}
\label{sec:ablation_studies}

We perform ablation studies to isolate and validate the contribution of each primary component of APN. The experiments, summarized in Table~\ref{tab:ablation_results}, yield three key observations. First, replacing the \textit{Query-Based Aggregation} with a basic linear layer leads to the most significant performance degradation. This demonstrates that a simple summarization is insufficient and that our sophisticated query mechanism is essential for dynamically identifying and prioritizing the most salient historical patterns for the forecasting task. Second, ablating the \textit{Weighted Aggregation} scheme and reverting to a simple average over hard-cut patches also results in a substantial decline in performance. This empirically confirms that our soft-aggregation design is critical for preventing information loss at patch boundaries and ensuring complete information coverage. Finally, removing the \textit{Adaptive Patching} mechanism in favor of fixed, randomly partitioned windows consistently diminishes performance, highlighting the value of learning dynamic, data-driven temporal boundaries over imposing rigid, arbitrary ones.

\begin{figure*}[t]
  \centering
  \subfloat[Patch Number] 
  {\includegraphics[width=0.247\textwidth]{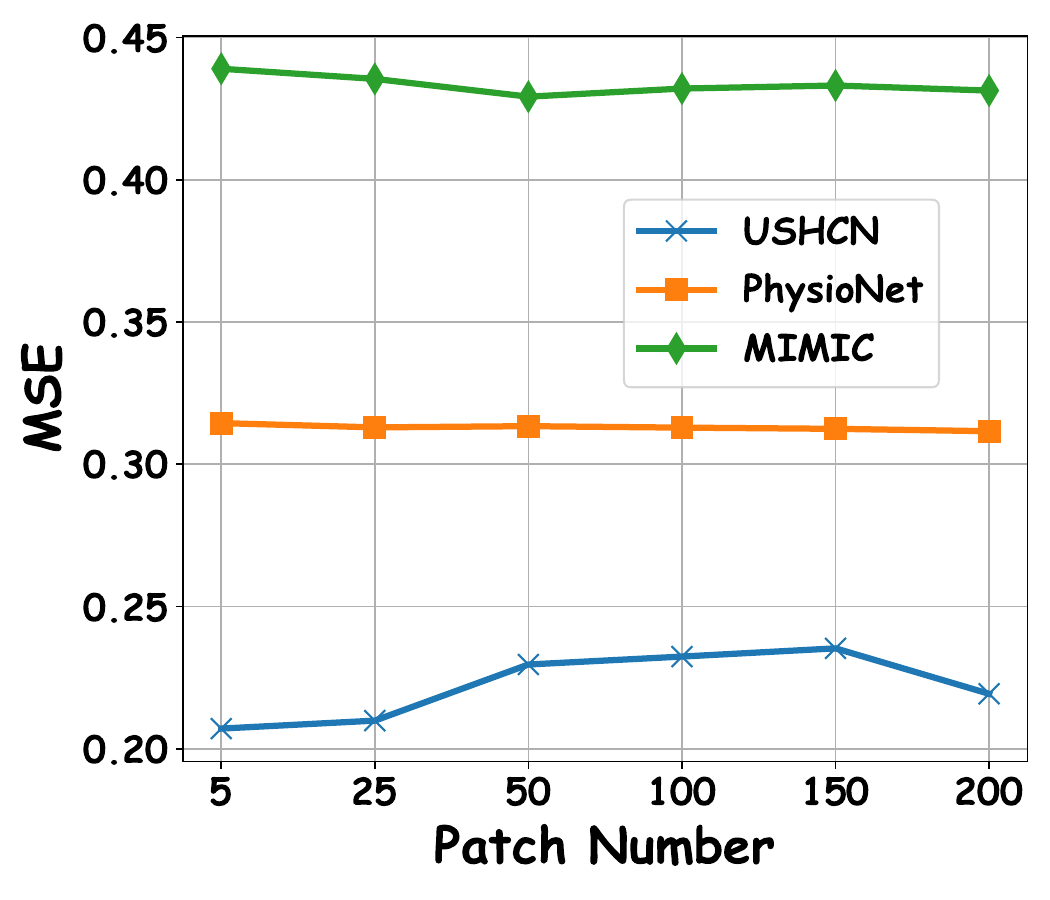}\label{fig:patch_number_pmu}} 
  \subfloat[Patch Number]
  {\includegraphics[width=0.247\textwidth]{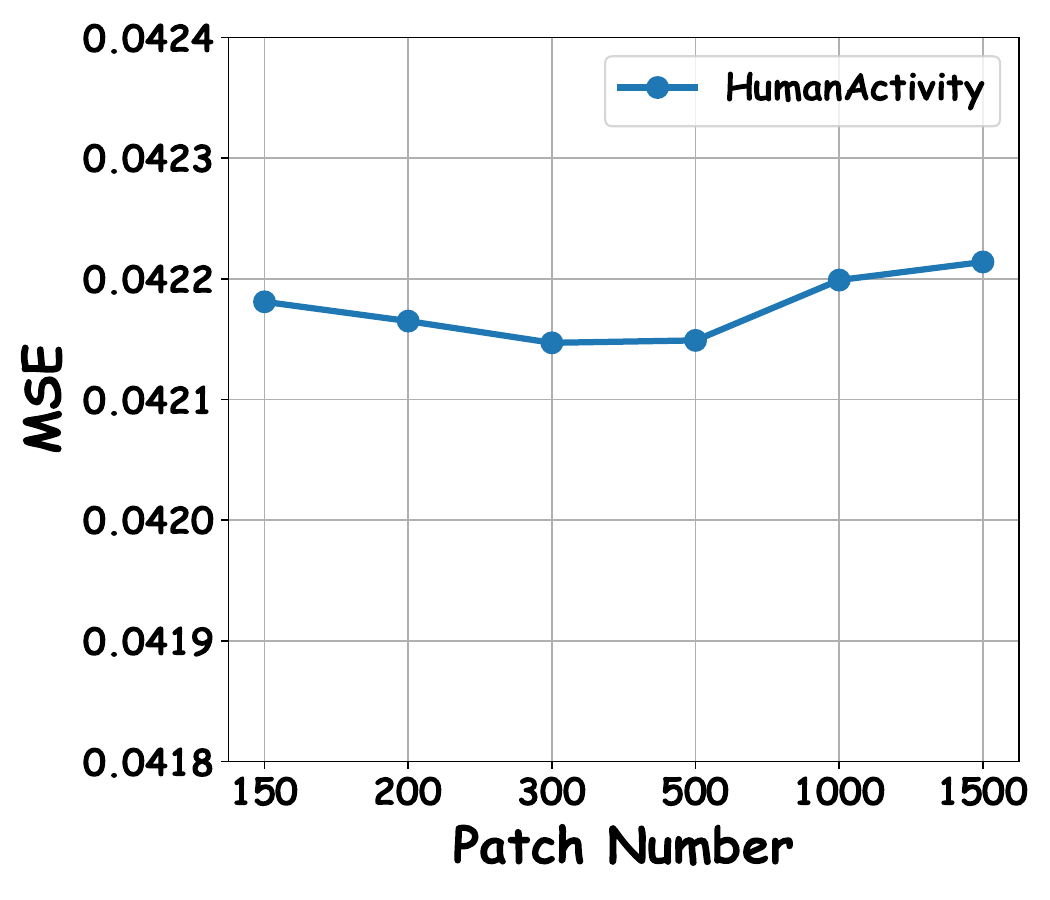}\label{fig:patch_number_activity}}
  \subfloat[Hidden Dimension]
  {\includegraphics[width=0.247\textwidth]{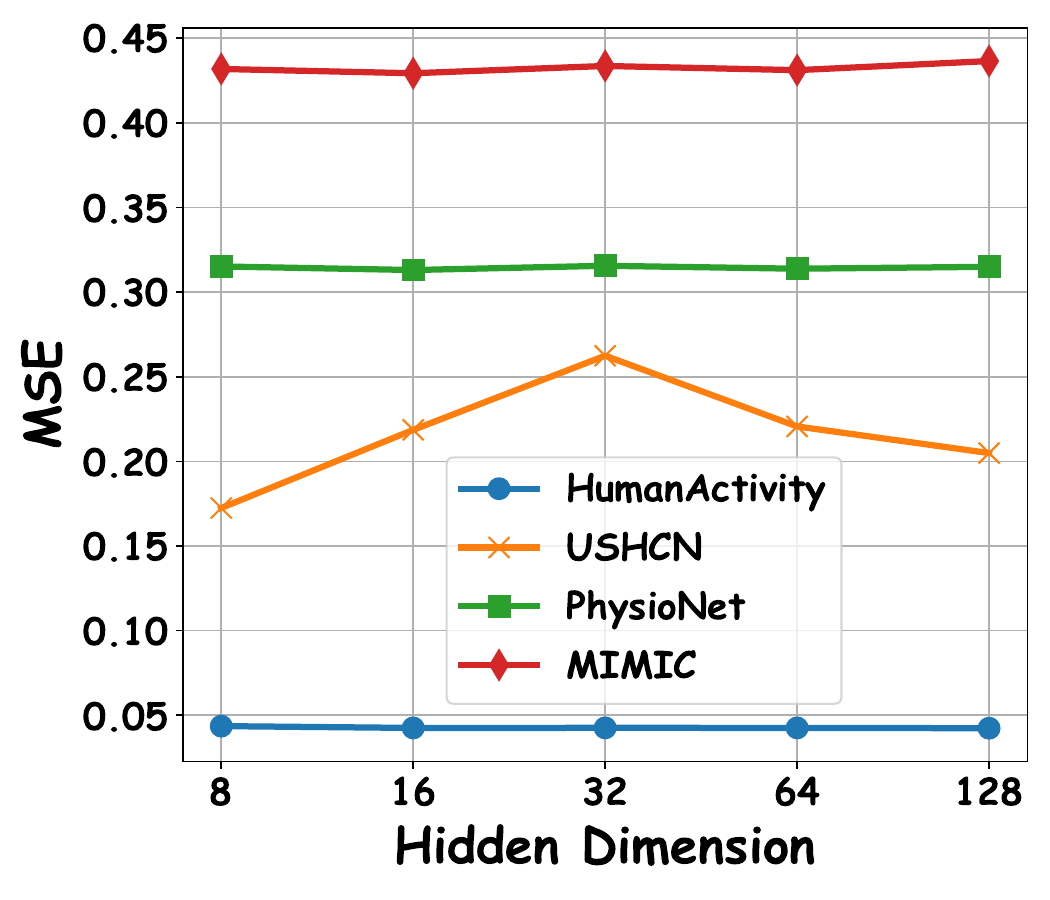}\label{fig:hidden_dimension_sensitivity}}
 \subfloat[Dimension of TE]
  {\includegraphics[width=0.247\textwidth]{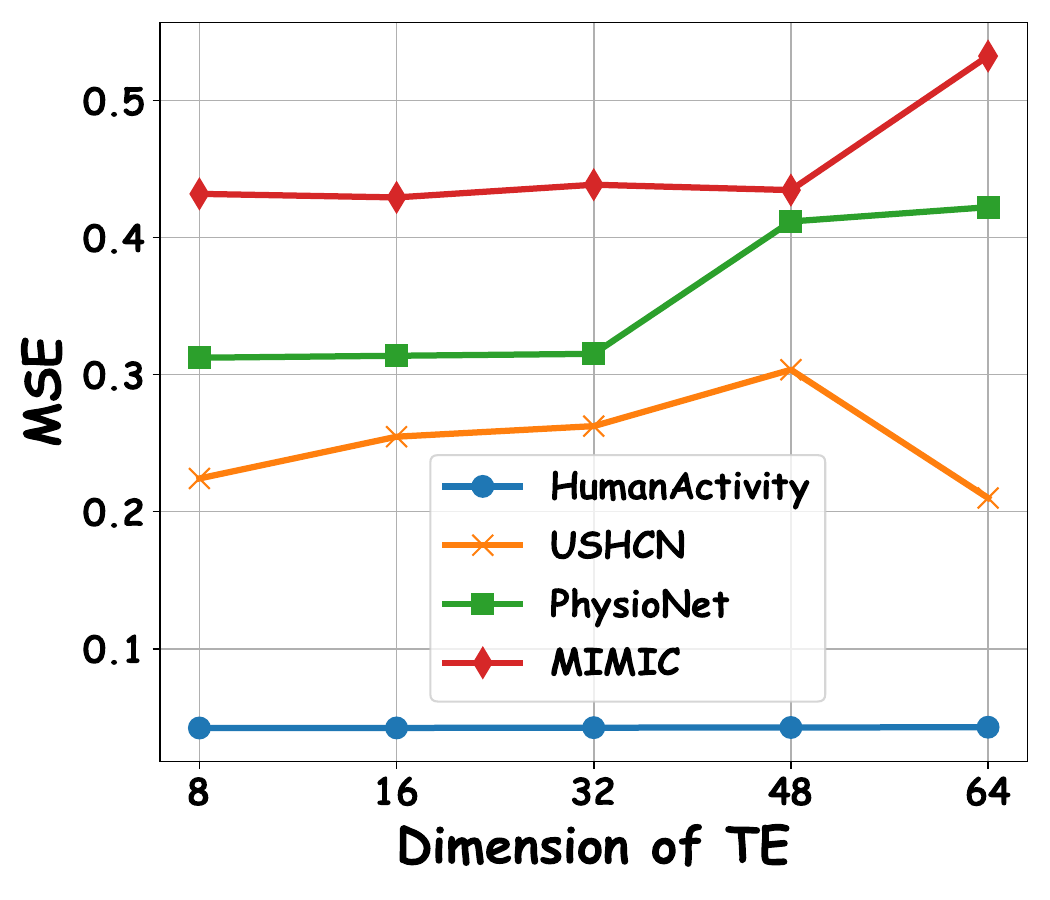}\label{fig:te_dim_sensitivity}} 
  \caption{Parameter sensitivity studies of main hyperparameters in APN.}
\label{fig:parameter_sensitivity}
\end{figure*}

\begin{figure}[t]
  \centering
  \subfloat[Peak GPU Memory (GB)\label{fig:peak_gpu_memory}]
  {\includegraphics[width=0.235\textwidth]{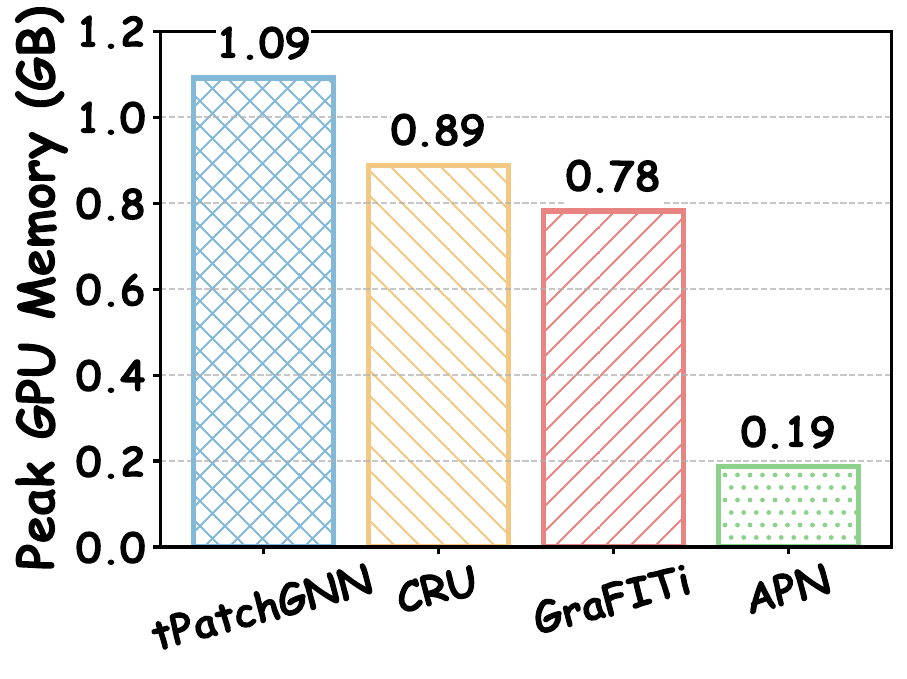}} 
  \hfill
  \subfloat[Parameters (M)\label{fig:num_parameters}]
  {\includegraphics[width=0.235\textwidth]{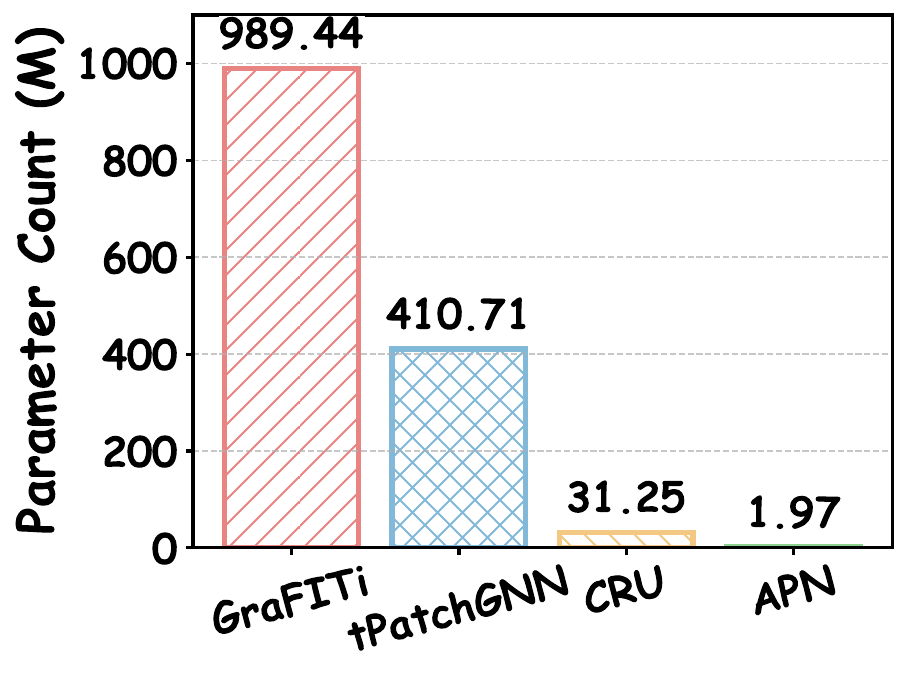}} 
  \\
  \subfloat[Training Step Time (ms)\label{fig:training_time}]
  {\includegraphics[width=0.235\textwidth]{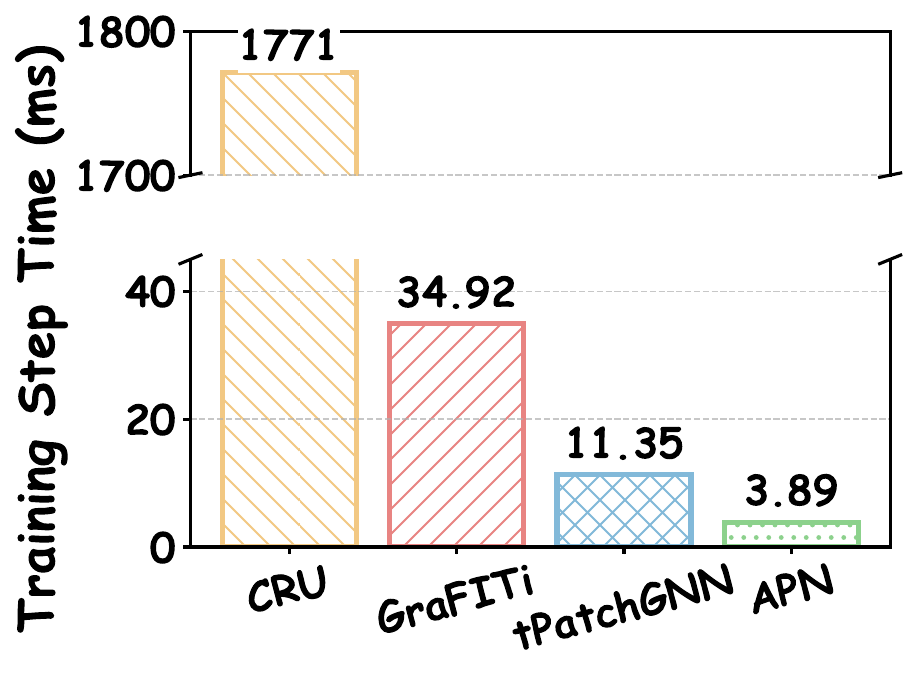}}
  \hfill
  \subfloat[Inference Step Time (ms)\label{fig:inference_time}]
  {\includegraphics[width=0.235\textwidth]{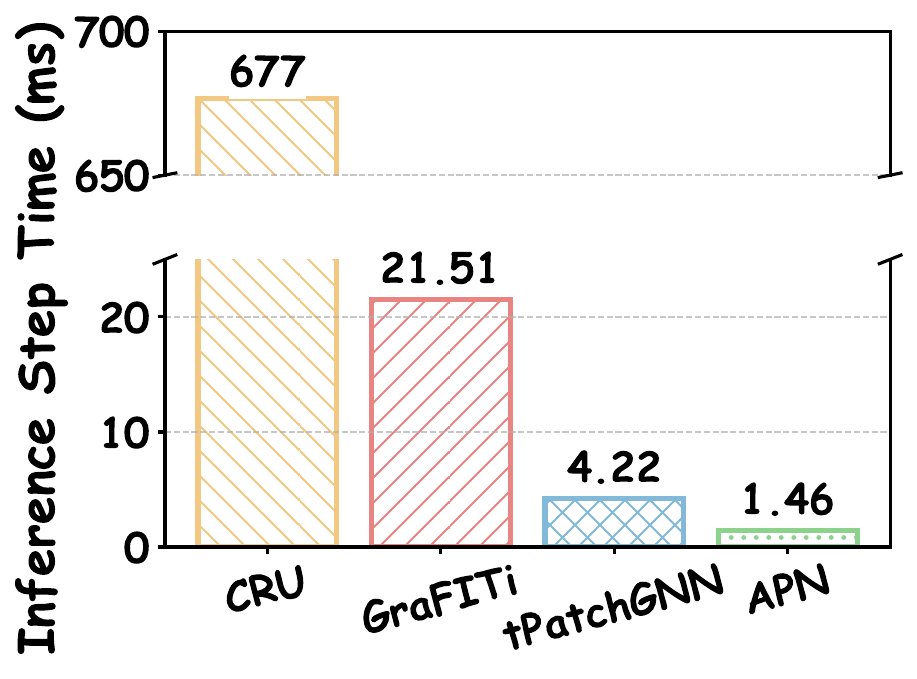}}
  
  \caption{Comparison of computational efficiency for APN and three representative baselines. We evaluate four key metrics: (a) peak GPU memory (GB) during a single training step, (b) total number of parameters (M), (c) average training time per step (ms), and (d) average inference time per step (ms). All experiments are conducted on the USHCN dataset with a consistent batch size of 32 to ensure a fair comparison. For all metrics, a lower value indicates better performance.}
  \label{fig:model_performance_comparison}
\end{figure}

\subsection{Parameter Sensitivity}
\label{sec:parameter_sensitivity}
To gain a deeper understanding of the impact of key hyperparameters on the performance of the APN model, we conduct parameter sensitivity analyses, focusing primarily on the number of patches ($P$), the model's hidden dimension ($D$), and the time encoding dimension ($D_{te}$)---see Figure~\ref{fig:parameter_sensitivity}. We have the following observations: 1) Figure~\ref{fig:hidden_dimension_sensitivity} reveals the impact of the model's hidden dimension ($D$). For most datasets, performance remains relatively stable across different dimensions, indicating that an excessively large dimension does not bring significant improvement. This suggests that a moderately sized $D$ is sufficient to capture effective information without unnecessary complexity. 2) Figure~\ref{fig:te_dim_sensitivity} shows the influence of the time encoding dimension ($D_{te}$). The model exhibits greater sensitivity to this parameter, as performance on datasets like PhysioNet and MIMIC degrades significantly when $D_{te}$ becomes too large. This indicates that a moderate $D_{te}$ is optimal for providing effective temporal information, while a higher dimension may introduce noise. 3) From Figures \ref{fig:patch_number_pmu} and \ref{fig:patch_number_activity}, we observe the effect of the patch number ($P$). For most datasets, model performance is not highly sensitive to the number of patches, suggesting that the model can robustly capture local features across a reasonable range of $P$ values. 4) The APN model exhibits a certain sensitivity to these core hyperparameters, especially $D_{te}$, but it generally achieves competitive results within a reasonable range. In practical applications, appropriate tuning should be performed based on the specific characteristics of the dataset.

\subsection{Scalability and Efficiency Analysis}
\label{sec:scalability_efficiency}

To comprehensively evaluate the potential of APN in practical deployment, we conduct a comparative analysis of computational efficiency and resource consumption between APN and three representative baseline models (GraFITi \citep{GraFITi}, CRU \citep{CRU}, tPatchGNN \citep{tPatchGNN}) on the USHCN dataset, with the batch size uniformly set to 32---see Figure~\ref{fig:model_performance_comparison}. The results demonstrate that APN exhibits significant advantages across all key efficiency metrics. This is attributed to its core TAPA module, which efficiently generates information-condensed and regularized initial representations. These representations support a highly streamlined information aggregation and prediction architecture, ultimately achieving a lightweight and highly efficient model design.

\section{Conclusion}
\label{sec:conclusion}
This paper addresses the challenges of IMTS forecasting by proposing a general and efficient framework, APN. The core of APN lies in its novel Time-Aware Patch Aggregation (TAPA) module. This module introduces an aggregation-based paradigm, distinct from methods that rely on fixed patching or interpolation. By learning dynamic boundaries and applying a time-aware soft-weighting strategy, TAPA directly aggregates information from all raw observations to generate high-quality, regularized patch representations. This design choice not only ensures full data coverage and robustly handles information density variations but also enables a streamlined and efficient overall architecture. Leveraging these superior representations, APN employs a concise query module and a shallow MLP to make predictions. Extensive experiments on multiple public IMTS benchmark datasets demonstrate that APN significantly outperforms existing state-of-the-art methods in both prediction accuracy and computational efficiency.

\section{Acknowledgments}
This work was partially supported by the National Natural Science Foundation of China (No.62472174) and the Fundamental Research Funds for the Central Universities.

\bibliography{aaai2026}
\end{document}